\documentclass[letterpaper]{article} % DO NOT CHANGE THIS
\usepackage{aaai2027}  % DO NOT CHANGE THIS
\nocopyright
\usepackage[hyphens]{url}  % DO NOT CHANGE THIS
\usepackage{graphicx} % DO NOT CHANGE THIS
\usepackage{natbib}  % DO NOT CHANGE THIS AND DO NOT ADD ANY OPTIONS TO IT
\usepackage{caption} % DO NOT CHANGE THIS AND DO NOT ADD ANY OPTIONS TO IT
\usepackage{algorithm}
\usepackage{algorithmic}
\usepackage{amsmath}
\usepackage{booktabs}
\usepackage{multirow}
\usepackage{amssymb}
\usepackage{adjustbox}
\usepackage{newfloat}
\usepackage{listings}
\DeclareCaptionStyle{ruled}{labelfont=normalfont,labelsep=colon,strut=off} % DO NOT CHANGE THIS
\floatstyle{ruled}
\newfloat{listing}{tb}{lst}{}
\floatname{listing}{Listing}

\usepackage{booktabs}

\title{OC-VLA++: Monocular Geometry-Guided Cross-View Consistency for Viewpoint-Robust Robotic Manipulation}

\author{
    Tianyi Zhang\textsuperscript{\rm 1} \ \ Ziyang Gong\textsuperscript{\rm 2} \ \ Zhenjie Yang\textsuperscript{\rm 3} \ \ Zhe Qian\textsuperscript{\rm 4} \ \ Haonan Duan\textsuperscript{\rm 5}\corresponding
}
\affiliations{
    \textsuperscript{\rm 1}Zhejiang University \ \
    \textsuperscript{\rm 2}Shanghai Jiao Tong University \ \
    \textsuperscript{\rm 3}The University of Hong Kong \\
    \textsuperscript{\rm 4}Renmin University of China \ \
    \textsuperscript{\rm 5}NVIDIA \\
   
}

\newcommand{\method}{OC-VLA++}

\newcommand{\robot}{\mathrm{robot}}

\begin{document}

\maketitle
\begin{figure*}[t]
\includegraphics[width=\linewidth]{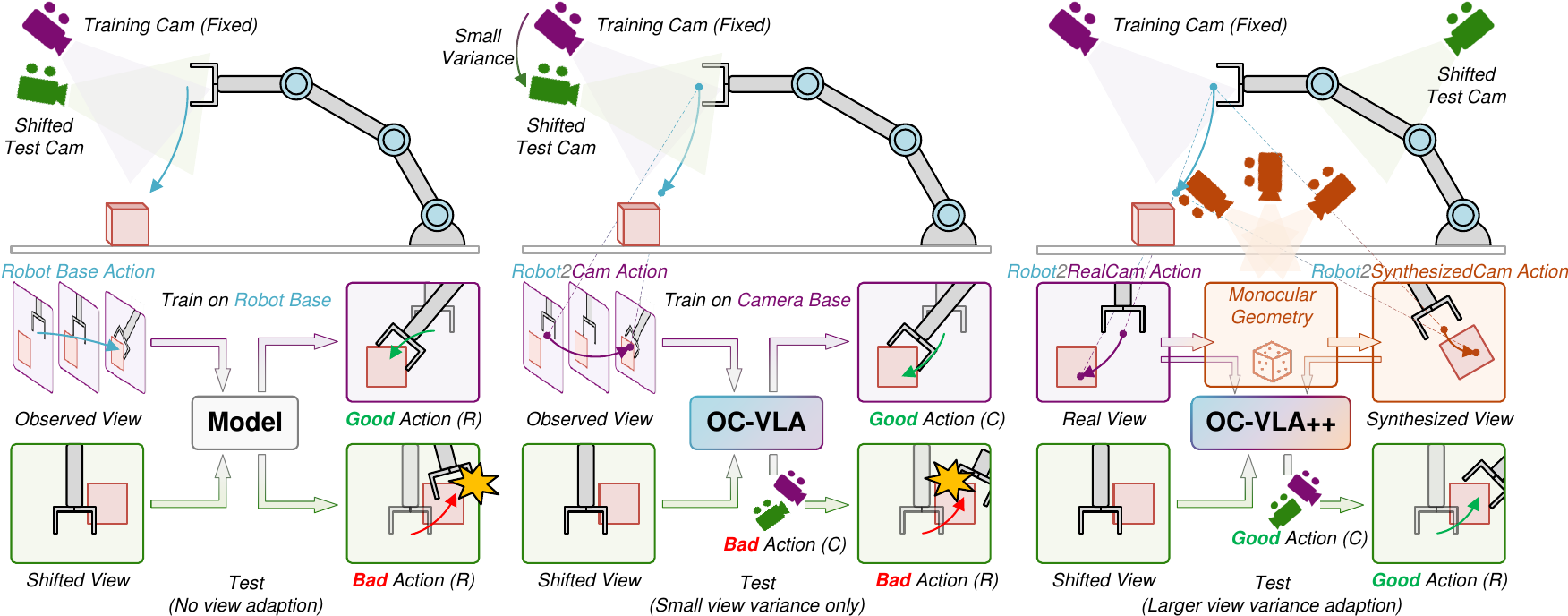}
% \fbox{\parbox[c][1.75in][c]{0.94\textwidth}{
\centering

\caption{Overview of OC-VLA++. Compared with OC-VLA \cite{OC-VLA} and other methods, OC-VLA++ constructs geometry-guided paired views
with view-specific camera-space action targets and enforces cross-view
action equivariance in a shared robot frame, enabling robust action
prediction under substantial camera-pose shifts.}
\label{fig:overview}
\end{figure*}

\begin{abstract}

We propose \textbf{OC-VLA++}, an extension of OC-VLA for viewpoint
generalization under limited camera coverage. While OC-VLA grounds
robot actions in the camera coordinate system to align action
supervision with visual observations, camera-space grounding alone can
still overfit to the few viewpoints observed during training.
OC-VLA++ addresses this limitation by introducing geometry-guided
paired-view supervision and an explicit cross-view action-equivariance
objective. Given paired observations of the same manipulation scene
from geometrically related viewpoints, the model is trained such that
their camera-space predictions correspond to the same robot-frame
action. This objective explicitly supervises how action predictions
should transform across viewpoints, rather than relying solely on
image-level augmentation. Experiments demonstrate substantial
improvements in unseen-view generalization under limited camera
coverage, with performance degrading more gracefully under increasing
camera displacement. These results establish cross-view action
equivariance as an effective complement to observation-centric action
grounding for robust real-world deployment.
\end{abstract}

\section{Introduction}

Vision-Language-Action (VLA) models have been increasingly adopted for robotic manipulation, where they predict executable actions from multimodal inputs such as visual observations, natural-language instructions, robot proprioceptive states, and other sensory signals \cite{dp,pi0, pi05, dtp, openvla, openvla-oft, gr00tn1, dita, rdt1b, vlaser}. With the continued growth and diversification of pretraining data, together with advances in training strategies, VLA models have achieved substantial improvements in manipulation capability, enabling robots to tackle increasingly complex tasks over longer temporal horizons \cite{xiaomirobotics1, lingbotvla2, rynnbrain11, abotm0, qwenvla, qwenrobotmanip, acebrain05, internvla15}. Nevertheless, viewpoint variation remains a persistent source of failure. A policy trained with a fixed third-person camera may degrade sharply when the camera is translated or rotated, even when the task, robot state, and object configuration remain unchanged. This issue is especially consequential in real deployments, where demonstrations are commonly collected from one or a few convenient camera placements rather than from a dense distribution of viewpoints.

One of the central challenges arises from the mismatch between the coordinate systems used to represent visual observations and action targets. Images are naturally organized in the camera frame, whereas end-effector actions are typically annotated in the robot base frame. OC-VLA \cite{OC-VLA} addresses this observation--action mismatch by transforming robot actions into the coordinate system of the observing camera and training the policy to predict camera-space actions. This observation-centric representation simplifies the mapping from image evidence to control and improves generalization across camera configurations without changing the underlying policy architecture.

However, the effectiveness of OC-VLA still depends on the viewpoint diversity available during training. With demonstrations collected from one or a few fixed cameras, the policy receives limited supervision for unseen viewpoints and still relies on camera-specific appearance and background cues. This limitation arises not from the observation-centric representation itself, but from insufficient cross-view supervision. 

To address this limitation, we propose OC-VLA++, which augments
observation-centric action grounding with geometry-guided paired-view
supervision and cross-view action equivariance. For each
training observation, we synthesize a nearby view with its
corresponding camera-space action target, and constrain predictions
from the paired views to represent the same physical action after
transformation into a shared robot frame. This provides explicit
geometric supervision across viewpoints, teaching the policy how
action predictions should transform with the camera pose rather than
merely exposing it to additional appearance variations through
conventional image augmentation. An overview is provided in Figure~\ref{fig:overview}.

\begin{figure*}[t]
\centering
\includegraphics[width=\linewidth]{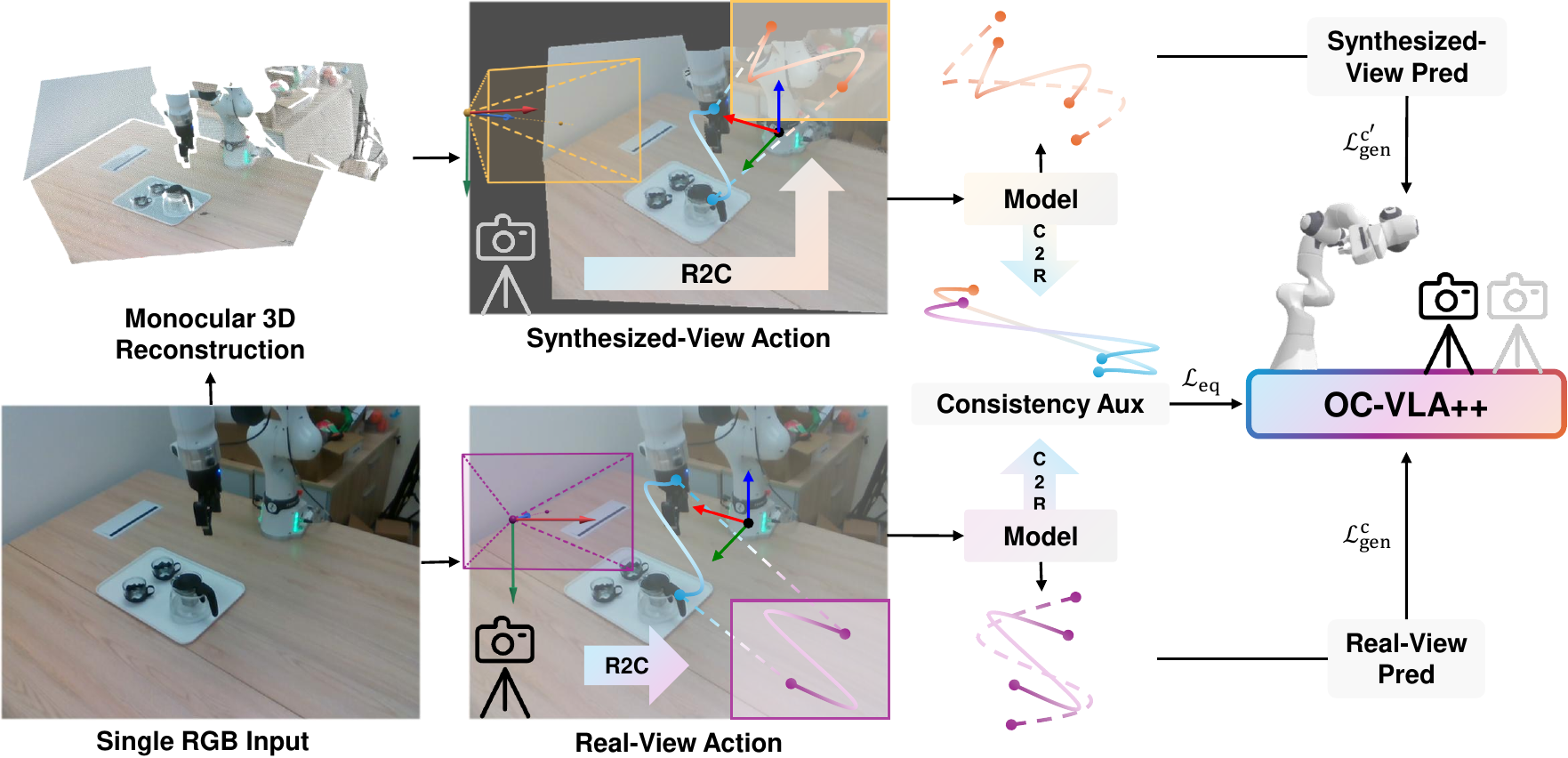}
% \fbox{\parbox[c][1.75in][c]{0.94\textwidth}{
\centering

\caption{Framework of OC-VLA++. Starting from a single RGB observation, a monocular geometry estimator
reconstructs a local 3D representation, which is transformed and
reprojected to synthesize a nearby view. The real and synthesized views are assigned
view-specific camera-space action targets and processed by the same
generative model. The recovered action estimates are transformed
into a shared robot frame and regularized by the cross-view
action-equivariance loss, while per-view generative losses supervise
both predictions. }
\label{fig:framework}
\end{figure*}

We evaluate OC-VLA++ in both real-robot and simulated environments,
focusing on viewpoint generalization when demonstrations are collected
from limited camera configurations. Across different policy families
and progressively larger camera-pose shifts, OC-VLA++ consistently
improves robustness over OC-VLA while preserving in-distribution
manipulation performance. Simulation experiments and component
ablations further confirm the complementary contributions of
geometry-guided paired-view supervision and cross-view action
equivariance. These results demonstrate that robust viewpoint
generalization requires not only camera-space action representations,
but also explicit supervision of how actions should transform across
views, highlighting its potential for viewpoint-robust manipulation.

\section{Related Work}

\subsection{Vision-Language-Action Policies}

VLA policies unify visual perception, language understanding, and action generation within a single framework. Early scalable policies demonstrated that heterogeneous robot data and pretrained vision-language representations can support broad manipulation capabilities \cite{rt1,rt2,octo,openvla}. Subsequent work introduced diffusion- and flow-based objectives for modeling continuous action sequences, improving the expressiveness and stability of action generation \cite{rdt1b,dita,pi0}. More recent studies have further explored efficient action representations, data-efficient adaptation, cross-embodiment transfer, and practical deployment \cite{openvla-oft,fast,pi05,gr00tn1,univla,smolvla,pi06}. Despite this progress, robustness to camera changes under limited viewpoint coverage remains comparatively underexplored. 

\subsection{Viewpoint-Robust Robot Learning}

Viewpoint robustness has been approached through explicit camera conditioning, geometry-aware visual representations, and camera-centric action formulations. Existing methods incorporate camera rays, calibrated geometric features, or robot-centric point representations to reduce the geometric ambiguity between perception and control \cite{knowcamera,g3vla,robotpointmap,roviaug,egodemogen,3ddiffact,gp3}. Some other works instead express actions relative to the observing camera, directly aligning the action representation with visual observations \cite{OC-VLA,camvla,equibot,etseed}. Another line of work synthesizes alternative viewpoints for training-time augmentation, test-time adaptation, or task-aware view selection \cite{vista,gensplat,robonvs,anycamvla,tavp,vistabot,sim2real,lift3d,activevla}. In contrast to methods that retain identical robot-frame targets or canonicalize observations to a reference view, \method{} explicitly regularizes the geometric relation between camera-space actions predicted from paired views of the same state.

\subsection{Monocular Geometry Estimation}

Monocular geometry estimation has advanced from generalizable relative-depth prediction to open-domain models capable of recovering metric depth and dense three-dimensional structure from a single image \cite{midas,dpt,depth_anything,unidepth,moge,moge2}. We use a frozen MoGe-2 model \cite{moge2} during offline preprocessing
to synthesize nearby novel views for real-world demonstrations. As
MoGe-2 is used only to construct paired training observations, it does
not alter the model architecture or incur inference-time overhead.

\section{Method}

\subsection{Overview}

OC-VLA++ augments observation-centric action grounding with
geometry-guided paired views and cross-view action
equivariance. During training, we pair each original observation with
a geometrically related view from a nearby camera pose
and supervise the two branches using action targets expressed in their
respective camera frames. We then recover action estimates from
both generative predictions, transform them into a shared robot frame,
and enforce consistency between the resulting robot-frame actions.
This design introduces explicit geometric supervision across views
without modifying the model architecture and applies uniformly to both diffusion based and
flow-matching based methods. The framework of OC-VLA++ is shown in Figure~\ref{fig:framework}.

\subsection{Geometry-Guided Paired-View Construction}

For each real-world observation \(I_t^c\), we synthesize nearby views
offline to construct geometrically related observation pairs. Given
the calibrated camera intrinsics \(K^c\), a frozen monocular geometry
estimator reconstructs a dense point map \(P_t^c\) and validity mask
\(M_t^c\):
\begin{equation}
\left(P_t^c,M_t^c\right)
=
f_{\mathrm{geo}}\left(I_t^c;K^c\right).
\end{equation}

We sample \(K\) local camera transformations
\(\{\Delta T_{c\rightarrow c_k}\}_{k=1}^{K}\). For each transformation,
the reconstructed points are rigidly transformed and reprojected onto
the target image plane:
\begin{equation}
\begin{aligned}
\left(\widetilde{I}_t^{c_k},M_t^{c_k}\right)
&=
\Pi\left(
P_t^c,I_t^c,M_t^c,K^c,
\Delta T_{c\rightarrow c_k}
\right),\\
T_{\robot\rightarrow c_k}
&=
\Delta T_{c\rightarrow c_k}
T_{\robot\rightarrow c},
\end{aligned}
\end{equation}
where \(\Pi\) denotes deterministic point transformation and
reprojection with depth-based visibility resolution.

The synthesized views, validity masks, and camera poses form an offline
view bank. During training, one synthesized view is paired with the
original observation, and its camera pose determines the corresponding
camera-space action target and cross-view transformation. We restrict
the transformations to a local neighborhood and discard views with
insufficient valid-pixel coverage. The entire view-generation process
is performed offline and is not used during inference.

In simulation, we directly render the same simulator state from a
sampled camera pose, providing exact paired observations without
monocular reconstruction errors.

Although the paired views correspond to the same robot-frame action,
their camera-space targets differ with the camera pose:
\begin{equation}
A_t^c=\Phi_c\left(A_t^{\robot}\right),
\qquad
A_t^{c'}=\Phi_{c'}\left(A_t^{\robot}\right),
\end{equation}
where \(\Phi_c\) maps a robot-frame action to camera \(c\). Each view
is therefore supervised in its own camera frame, rather than retaining
an unchanged action label as in standard image augmentation.

\subsection{Cross-View Action Equivariance}

Let \(c\) denote the original view and \(c'\) its paired synthesized
view. For \(v\in\{c,c'\}\), we denote the normalized camera-space
action target by \(A_t^v\) and the model condition by
\(
C_t^v=(I_t^v,\ell,s_t),
\)
which contains the visual observation, language instruction, and robot
state. The two branches share model parameters and use the same
generative timestep \(\tau\). Each branch retains the standard
generative objective of the underlying action model, denoted by
\(\mathcal{L}_{\mathrm{gen}}^v\).

The raw generative outputs of the paired branches do not directly
represent clean actions and are therefore unsuitable for action-level
consistency. We first recover a differentiable clean-action estimate
from each branch. Since the two estimates are expressed in different
camera frames, we then transform them into a shared robot frame before
imposing cross-view equivariance.

\paragraph{Clean-action recovery}
For diffusion-based models, the camera-space action target is perturbed
as
\begin{equation}
\begin{aligned}
X_\tau^v
&=
\alpha_\tau A_t^v+\sigma_\tau\epsilon^v,
\qquad
\epsilon^v\sim\mathcal{N}(0,I),\\
\widehat{\epsilon}^{\,v}
&=
\epsilon_\theta
\left(
X_\tau^v,\tau,C_t^v
\right).
\end{aligned}
\end{equation}
The corresponding clean-action estimate is recovered by
\begin{equation}
\widehat{A}_t^{\,v}
=
\frac{
X_\tau^v-\sigma_\tau\widehat{\epsilon}^{\,v}
}{
\alpha_\tau
}.
\label{eq:diffusion_recovery}
\end{equation}

For flow-matching models, we adopt the linear probability path
\begin{equation}
X_\tau^v
=
(1-\tau)Z^v+\tau A_t^v,
\qquad
Z^v\sim\mathcal{N}(0,I),
\end{equation}
with target velocity \(U^v=A_t^v-Z^v\). Given the predicted velocity
\(
\widehat{U}^{\,v}
=
u_\theta(X_\tau^v,\tau,C_t^v),
\)
the clean-action estimate is
\begin{equation}
\widehat{A}_t^{\,v}
=
X_\tau^v+(1-\tau)\widehat{U}^{\,v}.
\label{eq:flow_recovery}
\end{equation}
Both recovery operations are differentiable and are used only to
construct the cross-view objective; the original diffusion or
flow-matching training objective remains unchanged.

\paragraph{Robot-frame alignment}
We unnormalize the recovered actions and transform them from their
respective camera frames into the shared robot frame:
\begin{equation}
\begin{aligned}
\widehat{A}_t^{\robot,c}
&=
\operatorname{sg}
\left[
\Psi_c
\left(
\widehat{A}_t^{\,c}
\right)
\right],\\
\widehat{A}_t^{\robot,c'}
&=
\Psi_{c'}
\left(
\widehat{A}_t^{\,c'}
\right),
\end{aligned}
\label{eq:common_frame_action}
\end{equation}
where \(\Psi_v\) denotes action unnormalization followed by the
camera-to-robot coordinate transformation, applied independently to
each position in the action chunk.

We apply
\(\operatorname{sg}[\cdot]\) to the original-view prediction so that
the equivariance loss updates the shared model only through the
synthesized-view path. The original-view prediction therefore serves
as the reference for cross-view alignment, reducing the influence of
reconstruction artifacts in the synthesized observation. This
stop-gradient operation affects only the equivariance objective; the
original view remains fully supervised by
\(\mathcal{L}_{\mathrm{gen}}^c\).

\paragraph{Action-level equivariance loss.}
For each valid action position \(i\in\mathcal{I}\), we decompose the
robot-frame prediction into translation \(p_i^v\), unit quaternion
\(q_i^v\), and gripper command \(g_i^v\). We measure the rotational
discrepancy using the quaternion geodesic angle
\begin{equation}
\theta_i
=
2\arccos
\left(
\operatorname{clip}
\left(
\left|
\left\langle q_i^c,q_i^{c'}\right\rangle
\right|,
0,1
\right)
\right),
\label{eq:rotation_angle}
\end{equation}

The component-wise discrepancies are defined as
\begin{equation}
\begin{aligned}
d_p^i
&=
\min
\left(
\frac{1}{D_p}
\sum_{j=1}^{D_p}
\rho_{\beta_p}
\left(
p_{i,j}^c-p_{i,j}^{c'}
\right),
\kappa_p
\right),\\
d_R^i
&=
\min
\left(
\rho_{\beta_R}(\theta_i),
\kappa_R
\right),\\
d_g^i
&=
\rho_{\beta_g}
\left(
g_i^c-g_i^{c'}
\right),
\end{aligned}
\label{eq:component_discrepancies}
\end{equation}
where \(\rho_\beta\) denotes the Huber loss, \(D_p\) is the
dimensionality of the translation component, and \(\kappa_p\) and
\(\kappa_R\) limit the influence of large translation and rotation
discrepancies that may arise from synthesis artifacts.

The cross-view action-equivariance loss is
\begin{equation}
\mathcal{L}_{\mathrm{eq}}
=
\frac{1}{|\mathcal{I}|}
\sum_{i\in\mathcal{I}}
\left(
\lambda_p d_p^i
+
\lambda_R d_R^i
+
\lambda_g d_g^i
\right),
\label{eq:equivariance_loss}
\end{equation}
where \(\lambda_p\), \(\lambda_R\), and \(\lambda_g\) balance the
translation, rotation, and gripper components.

\subsection{Training Objective}

Let \(\mathcal{L}_{\mathrm{gen}}^v\) denote the standard generative
objective for view \(v\). The overall
training objective is
\begin{equation}
\mathcal{L}
=
\mathcal{L}_{\mathrm{gen}}^c
+
\lambda_{\mathrm{syn}}
\mathcal{L}_{\mathrm{gen}}^{c'}
+
\lambda_{\mathrm{eq}}
\mathcal{L}_{\mathrm{eq}},
\label{eq:total_loss}
\end{equation}
where \(\lambda_{\mathrm{syn}}\) weights the generative supervision on
the synthesized view, and \(\lambda_{\mathrm{eq}}\) controls the
cross-view action-equivariance objective. The first two terms supervise
each observation with its corresponding camera-space action target,
while the third aligns their recovered predictions in the shared robot
frame.

\section{Experiments}

We evaluate OC-VLA++ primarily on real-robot manipulation tasks,
complemented by simulation experiments. Our evaluation
addresses three questions: (1) whether OC-VLA++ preserves performance
under the training camera configuration, (2) whether it improves
robustness under progressively larger camera-pose shifts, and
(3) how geometry-guided paired-view supervision and cross-view action
equivariance contribute to the overall improvement.

To enable direct comparison with OC-VLA \cite{OC-VLA}, we follow its task
definitions, datasets, training protocols, and evaluation criteria as
closely as possible. We retain its original model architecture and
additionally evaluate a structurally different model to examine whether
the proposed cross-view supervision generalizes beyond the original
setting. The real-robot experiments measure both fixed-view performance
and robustness to camera displacement, while simulation provides exact
paired observations for controlled analysis and component ablations.

\subsection{Model Architectures}

We instantiate OC-VLA++ on two complementary model families. To
preserve direct comparability with OC-VLA \cite{OC-VLA}, we retain
Dita \cite{dita}, which directly denoises continuous action chunks
with a diffusion transformer without relying on a pretrained VLM
backbone. We additionally develop a Qwen3-VL-2B-based model
 \cite{qwen3vl} following the action-expert formulation of
\(\pi_0\) \cite{pi0}, where a pretrained VLM encodes the multimodal
context and a dedicated action expert generates continuous actions
through flow matching.

Following OC-VLA, we first pretrain Dita on DROID \cite{droid} before
adapting it to each target dataset. This stage provides diverse
robot-manipulation priors across scenes, tasks, and camera viewpoints
that are not inherited from a pretrained VLM backbone. In contrast,
the Qwen-based model already benefits from large-scale,
image-conditioned vision-language pretraining; we therefore fine-tune
it directly on the corresponding real-robot or simulation dataset
without an intermediate pretraining stage. Together, the two model families span different architectures,
pretraining regimes, and generative formulations, providing a broader
evaluation of OC-VLA++.

\subsection{Real-Robot Experiments}

\subsubsection{Setup and Data}

We conduct real-robot experiments using a 7-DoF Franka Emika Panda
manipulator equipped with a Robotiq 2F-85 gripper. Two calibrated Intel
RealSense D435i RGB-D cameras are used. One camera remains fixed for
demonstration collection and fixed-view evaluation, following a setup
comparable to OC-VLA \cite{OC-VLA}. The other camera is movable
during evaluation to measure robustness under camera-pose changes. The
dataset covers a diverse set of manipulation skills, including
pick-and-place, pouring, stacking, deformable-object manipulation,
pushing and pulling, and long-horizon interactions.

All training demonstrations are collected from the single fixed
camera, representing the limited-viewpoint coverage commonly
encountered in real-world data collection. Following OC-VLA and
Dita \cite{OC-VLA,dita}, each model is fine-tuned using 10
demonstrations per task. During camera-displacement evaluation, the
movable camera is placed at a sequence of unseen poses with
progressively larger translation and rotation relative to the training
view. We evaluate every method using 20 rollouts per task under each
camera configuration, increasing the 10-rollout protocol used in the
prior works to obtain a more reliable estimate of task success.

\begin{figure}
\centering

\includegraphics[width=\linewidth]{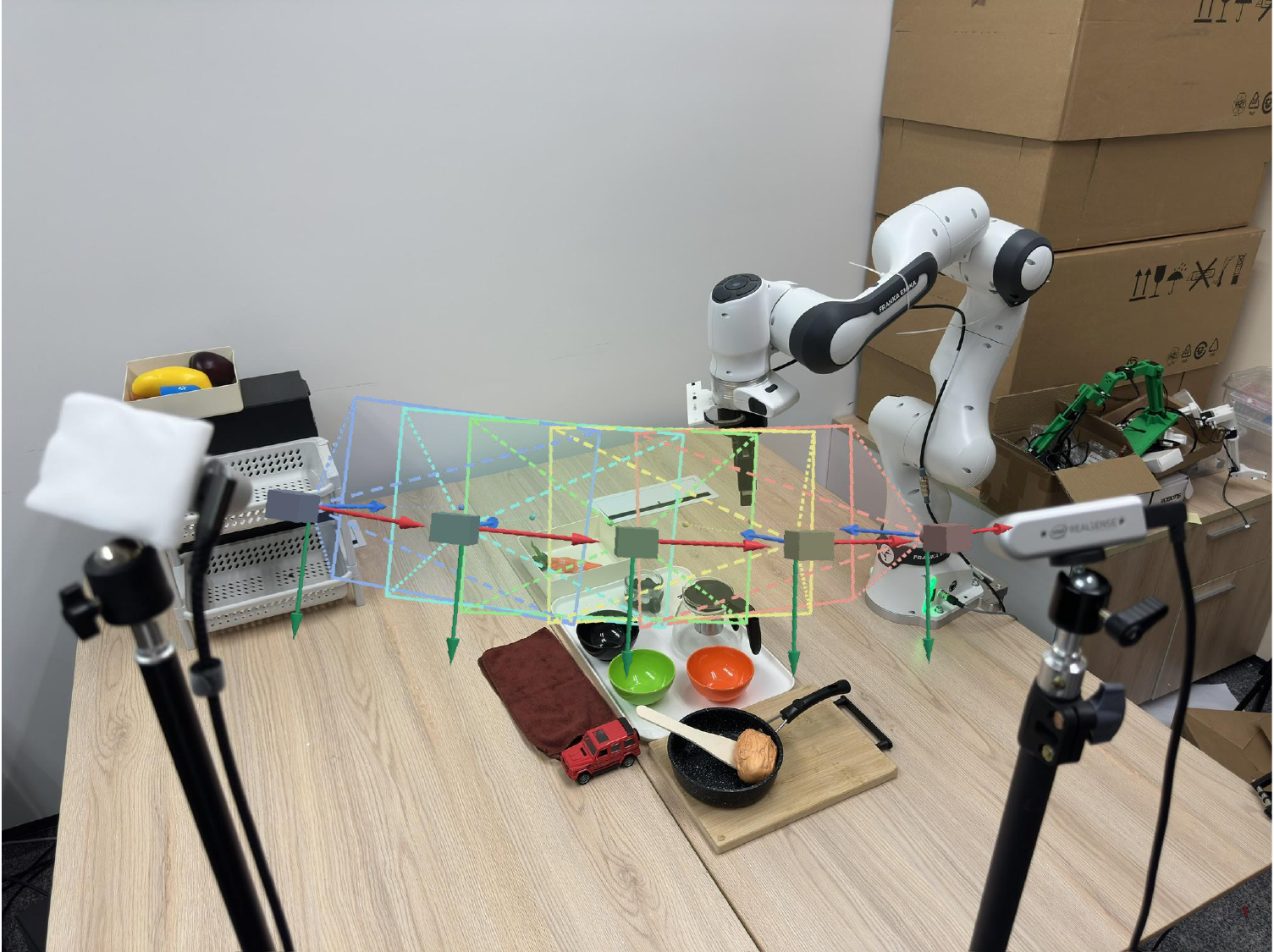}
\centering

\caption{Real-robot experimental setup. We use a Franka Emika Panda robot
equipped with a Robotiq 2F-85 gripper to execute the actions.
Two Intel RealSense D435i RGB-D cameras are used: the left camera
remains fixed for data collection and fixed-view evaluation while
the right camera is movable to evaluate robustness under camera
displacement.}
\label{fig:setup}
\end{figure}

\subsubsection{Synthesized View Pool Configuration}

For each recorded real-robot training frame, we construct an offline
pool of 10 synthesized views using MoGe-2 \cite{moge2}. Each view is
generated from a sampled local camera perturbation
\(\Delta T=(\Delta R,\Delta\mathbf{t})\), with
\(\lVert\Delta\mathbf{t}\rVert_2\leq15\,\mathrm{cm}\) and
\(d_{\mathrm{SO}(3)}(\Delta R,I)\leq10^\circ\).
The target-camera extrinsic is obtained by composing \(\Delta T\) with
the calibrated source-camera pose. During training, one synthesized
view is randomly sampled and paired with the original observation.
The view-generation pipeline is used only for offline preprocessing
and introduces no inference-time overhead.

\subsubsection{Fixed-View Evaluation}

We first evaluate all models using the fixed camera configuration
employed for data collection. This in-distribution setting examines
whether the additional paired-view supervision preserves manipulation
performance at the training viewpoint. We evaluate six tasks covering diverse interaction patterns.

\begin{table*}[!htbp]
\centering
\begin{tabular}{l|c|cccccc}
\toprule
{\bf Method}
& {\bf Avg}
& {\bf Pick Carrot}
& {\bf Pour Water}
& {\bf Fold Towel}
& {\bf Stack Bowls}
& {\bf Push Cars}
& {\bf Cook Potato} \\
\midrule

\multicolumn{8}{l}{\textit{External VLA Baselines}} \\
\(\pi_0\)
& 51.7\% & 70.0\% & 35.0\% & 60.0\% & 30.0\% & 55.0\% & 60.0\% \\
OpenVLA-OFT
& 63.3\% & 90.0\% & 50.0\% & 55.0\% & 30.0\% & 80.0\% & 75.0\% \\
\midrule

\multicolumn{8}{l}{\textit{DiT-Based Models}} \\
Dita
& 61.7\% & 75.0\% & 55.0\% & 85.0\% & 45.0\% & 50.0\% & 60.0\% \\
OC-VLA (Dita)
& 68.3\% & 70.0\% & 60.0\% & 95.0\% & 40.0\% & 70.0\% & 75.0\% \\
OC-VLA++ (Dita)
& 68.3\% & 80.0\% & 50.0\% & 95.0\% & 50.0\% & 60.0\% & 75.0\% \\
\midrule

\multicolumn{8}{l}{\textit{VLM--Action Expert Models}} \\
Qwen3-VL-2B
& 52.5\% & 70.0\% & 40.0\% & 65.0\% & 30.0\% & 50.0\% & 60.0\% \\
OC-VLA (Qwen3-VL-2B)
& 60.0\% & 75.0\% & 50.0\% & 80.0\% & 35.0\% & 60.0\% & 60.0\% \\
OC-VLA++ (Qwen3-VL-2B)
& 59.2\% & 75.0\% & 45.0\% & 75.0\% & 45.0\% & 55.0\% & 60.0\% \\
\bottomrule
\end{tabular}

\caption{
Real-robot evaluation at the fixed training view. Each model is
evaluated over 20 rollouts per task. Qwen3-VL-2B denotes our model that
combines a pretrained Qwen3-VL-2B vision-language backbone \cite{qwen3vl} with a
dedicated action expert for continuous action generation through flow
matching, following the general architecture of
\(\pi_0\) \cite{pi0}.
}
\label{tab:realrobotfixcam}
\end{table*}

As shown in Table~\ref{tab:realrobotfixcam}, OC-VLA++ achieves average
success rates of 68.3\% and 59.2\% with the Dita \cite{dita} and Qwen-based
architectures \cite{qwen3vl}, respectively. These results are nearly identical to
those of the corresponding OC-VLA variants, which achieve 68.3\% and
60.0\%. Thus, geometry-guided paired-view supervision and cross-view
action equivariance preserve performance at the training viewpoint.
Their primary benefit emerges under viewpoint shifts beyond the
training distribution, as evaluated next.

\subsubsection{Camera-Displacement Evaluation}

\begin{figure*}
    \centering
    \includegraphics[
        width=\linewidth
    ]{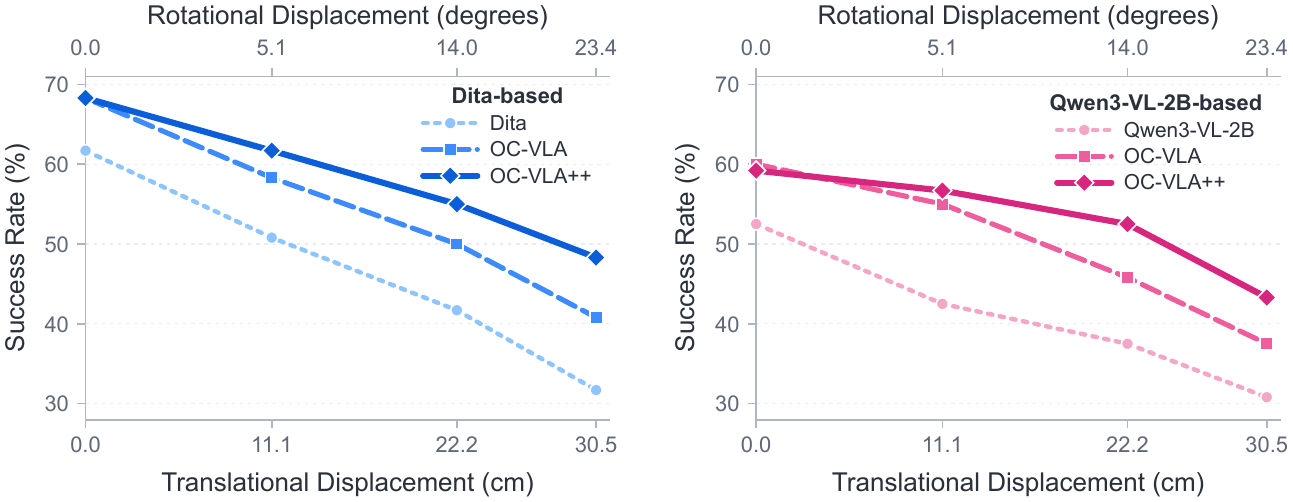}
    \caption{
        Real-robot success rates under increasing camera displacement.
        The left and right panels show the Dita-based and Qwen3-VL-2B-based
        policies, respectively. The lower and upper
        horizontal axes indicate the translational and rotational offsets
        of the same camera poses from the data-collection view.
    }
    \label{fig:camera_displacement_all_methods}
\end{figure*}

We next evaluate viewpoint generalization by progressively moving the
observation camera away from the pose used for demonstration
collection. Starting from the training camera configuration, we evaluate
four camera poses with progressively larger viewpoint shifts.
For each pose, translational change is measured by the
Euclidean norm of the 3D displacement, while rotational
change is measured by the geodesic angle relative to the
training orientation. All models are evaluated
at the same poses under an identical rollout protocol. The average success rates across the six tasks
are reported in Figure~\ref{fig:camera_displacement_all_methods}.

As the camera moves farther from the training configuration, the
performance of all models decreases. However, OC-VLA++ exhibits consistently slower degradation with both
the Dita and Qwen3-VL-2B-based architectures. At the largest camera
shift, OC-VLA++ achieves success rates of 48.3\% and 43.3\%,
respectively, compared with 40.8\% and 37.5\% achieved by the
corresponding OC-VLA variants. The corresponding robot-base-frame models obtain
only \(31.7\%\) and \(30.8\%\). These results show that
geometry-guided paired-view supervision and cross-view action
equivariance improve robustness to camera-pose changes while
preserving performance at the training viewpoint.

\begin{figure*}
    \centering
    \includegraphics[
        width=\linewidth
    ]{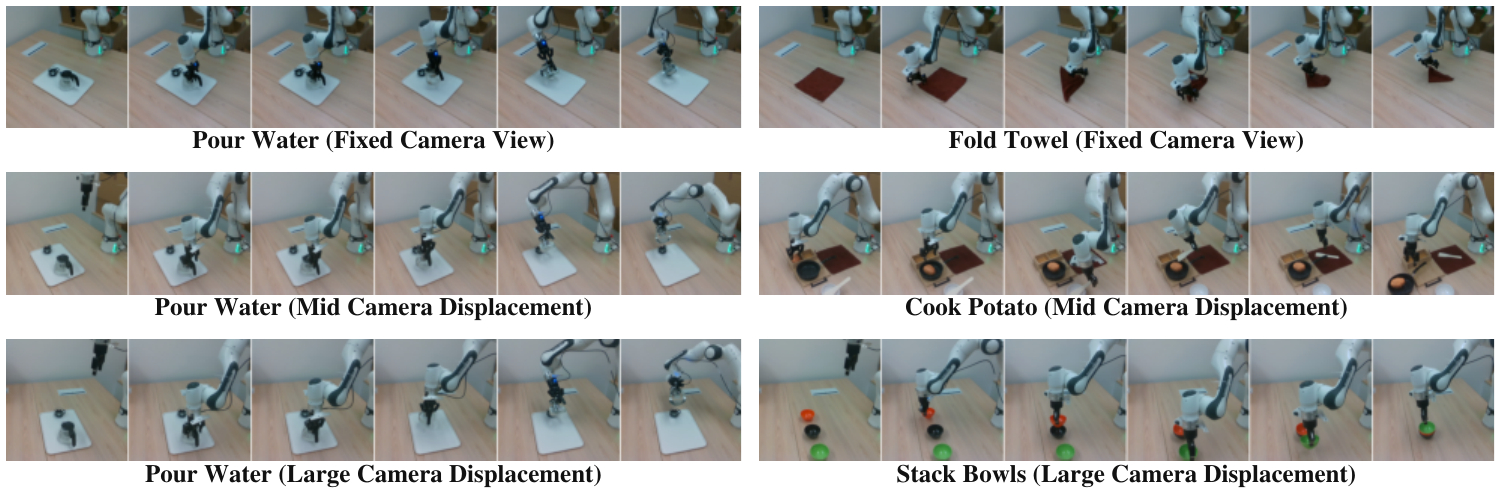}
        \caption{
Qualitative real-robot results of OC-VLA++ under increasing camera
displacement from top to bottom.
}
    
    \label{fig:real_robot_qualitative_results}
\end{figure*}

Notably, the two most displaced evaluation poses exceed the local
perturbation range used to construct the synthesized-view pools
(\(15\,\mathrm{cm}\) and \(10^\circ\)). The persistent gains at these
poses indicate that locally generated paired-view supervision can
generalize to substantially larger physical camera shifts. Qualitative
results are provided in
Figure~\ref{fig:real_robot_qualitative_results}.

\begin{table*}[!htbp]

\centering

\begin{tabular}{c|cccccc}

{\bf Method}   & {\bf Average} & { \bf PickC} & {\bf StackC} & {\bf SingleYCB} & {\bf ClutterYCB} & {\bf SingleEGAD}\\

\midrule

Dita &    38.6\%  &  61.0\% & 51.0\%  & 28.0\%  & 8.0\%  & 45.0\%  \\

OC-VLA (Dita) &   52.4\% & 80.0\%  &  65.0\% & {\bf 48.0\%}  & 19.0\%  & 50.0\%  \\

OC-VLA++ (Dita) &   {\bf 56.8\%} & {\bf 82.0\%}  &  {\bf 77.0\%} & {\bf 48.0\%}  & {\bf 21.0\%}  & {\bf 56.0\%}  \\

\end{tabular}

\caption{Comparison on ManiSkill2 \cite{maniskill2} in terms of task
success rate.  SingleYCB indicates PickSingleYCB, ClutterYCB indicates PickClutterYCB, SingleEGAD indicates PickSingleEGAD. The results of Dita and OC-VLA(Dita) are from OC-VLA \cite{OC-VLA}.}

\label{tab:maniskill_main}

\end{table*}
\subsection{Simulation Experiments}
\subsubsection{Setup and Data}

We complement the real-robot evaluation with controlled simulation
experiments, where camera poses and environment states can be varied
precisely and paired observations can be rendered from identical
states. This setting enables large-scale multi-view evaluation and
controlled ablations that would be costly to conduct on physical
hardware.

We follow the ManiSkill2-based protocol of
OC-VLA \cite{maniskill2,OC-VLA} and use the same five tasks:
\textit{PickCube}, \textit{StackCube}, \textit{PickSingleYCB},
\textit{PickClutterYCB}, and \textit{PickSingleEGAD}. For each
trajectory, we sample 20 camera poses from the same pool of 300K
candidate poses and render the corresponding observations. The task
definitions, camera-sampling procedure, and data scale are kept
consistent with OC-VLA to enable direct comparison.

To construct paired views, we re-render the same simulator state from
an additional sampled camera pose. Unlike the monocular
geometry-based procedure used for real-robot data, this process
provides exact observations and camera transformations without
reconstruction artifacts, allowing us to evaluate cross-view
supervision under fully controlled conditions.

\subsubsection{Evaluation under Multi-View Training}

We evaluate OC-VLA++ under the multi-view training protocol of
OC-VLA. All methods use the same Dita architecture, task set, training
trajectories, and camera-sampling procedure. Each trajectory is rendered
from 20 camera poses sampled from the same pool of 300K candidates,
providing broad viewpoint coverage during training. This setting
therefore isolates the contribution of the proposed cross-view
supervision when diverse camera observations are already available.

As shown in Table~\ref{tab:maniskill_main}, observation-centric action
grounding improves the average success rate from \(38.6\%\) with Dita
to \(52.4\%\) with OC-VLA. OC-VLA++ further raises the average success
rate to \(56.8\%\), yielding  a 4.4-percentage-point improvement over
OC-VLA. The smaller margin compared with the limited-view real-robot
setting is consistent with the broad camera coverage already provided
by multi-view training. 

Importantly, OC-VLA++ does not merely expose the model to additional
views. It explicitly constrains predictions from paired observations
to correspond to the same robot-frame action. The improvement over
OC-VLA therefore shows that cross-view action equivariance remains
beneficial even when the training data already contain substantial
viewpoint diversity.
\subsubsection{Ablation Study}

We ablate the two components introduced on top of OC-VLA.
\textit{OC-VLA + Synth.} adds generative supervision on the
synthesized view, whereas \textit{OC-VLA + Equiv.} uses it
only to compute the cross-view action equivariance loss. OC-VLA++ combines both components.

\begin{table}[t]
    \centering
    \setlength{\tabcolsep}{4.5pt}
    \begin{tabular}{l|ccc|c}
        \toprule
        \multirow{2}{*}{\textbf{Method}}
        & \multicolumn{3}{c|}{\textbf{Training Objectives}}
        & \multirow{2}{*}{\textbf{Avg.}} \\
        \cmidrule(lr){2-4}
        &
        \(\boldsymbol{\mathcal{L}_{\mathrm{gen}}^{c}}\)
        &
        \(\boldsymbol{\mathcal{L}_{\mathrm{gen}}^{c'}}\)
        &
        \(\boldsymbol{\mathcal{L}_{\mathrm{eq}}}\)
        & \\
        \midrule

        OC-VLA (Dita)
        & \checkmark
        & --
        & --
        & 52.4\% \\

        OC-VLA (Dita) + Synth.
        & \checkmark
        & \checkmark
        & --
        & 53.4\% \\

        OC-VLA (Dita) + Equiv.
        & \checkmark
        & --
        & \checkmark
        & 54.2\% \\

        OC-VLA++ (Dita)
        & \checkmark
        & \checkmark
        & \checkmark
        & \textbf{56.8\%} \\

        \bottomrule
    \end{tabular}
    \caption{
        Ablation study on ManiSkill2 \cite{maniskill2}. The generative objectives
        \(\mathcal{L}_{\mathrm{gen}}^{c}\) and
        \(\mathcal{L}_{\mathrm{gen}}^{c'}\) supervise the original and
        synthesized views, respectively, while
        \(\mathcal{L}_{\mathrm{eq}}\) enforces cross-view action
        equivariance. All variants use the same architecture,
        training data, and evaluation protocol.
    }
    \label{tab:maniskill_ablation}
\end{table}

As shown in Table~\ref{tab:maniskill_ablation}, adding generative
supervision on the synthesized view improves the average success rate
from \(52.4\%\) to \(53.4\%\). Cross-view action equivariance alone
achieves \(54.2\%\), showing that explicitly aligning paired
predictions in the robot frame provides benefits beyond additional
view supervision. Combining both components yields the best result of
\(56.8\%\), demonstrating that synthesized-view supervision and cross-view action
equivariance provide complementary training signals.

\section{Conclusions}

We introduced OC-VLA++, which complements observation centric action
grounding with geometry-guided paired-view supervision and cross-view
action equivariance. By aligning predictions from geometrically
related observations in a shared robot frame, OC-VLA++ largely
preserves performance at the training viewpoint while exhibiting more
graceful degradation under increasing camera displacement. Real-robot
and simulation experiments demonstrate its effectiveness across
different model architectures and camera-coverage regimes, without
requiring geometry estimation or view synthesis at inference time. Our results establish action-level cross-view supervision as an
effective complement to camera-space action grounding. By explicitly
modeling how action predictions should transform across viewpoints,
OC-VLA++ reduces sensitivity to camera configurations and provides a
practical approach to viewpoint-robust manipulation under limited-view
data collection.

\clearpage \clearpage \clearpage \clearpage
\bibliography{aaai2027}
\clearpage \clearpage \clearpage \clearpage
\appendix
\section{Appendix}
\subsection{Model Structures and Details}
\begin{figure*}[!htbp]
\centering
\includegraphics[width=\linewidth]{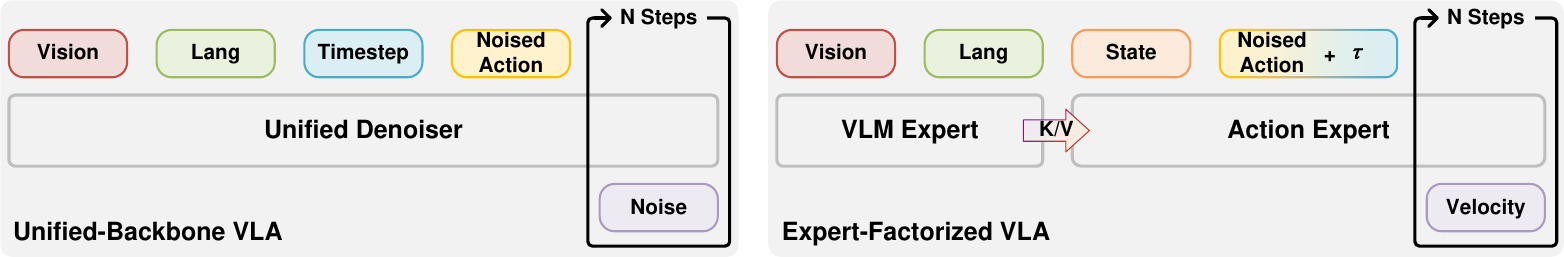}
% \fbox{\parbox[c][1.75in][c]{0.94\textwidth}{
\centering

\caption{
Model structures used in our experiments. The Dita-based model \cite{dita}
(left) uses a unified denoiser to predict action noise, whereas the
Qwen3-VL-2B-based model (right) \cite{qwen3vl} uses a VLM backbone and an
action expert to predict flow velocity.
}
\label{fig:model_struct}
\end{figure*}
OC-VLA++ is architecture-agnostic and is instantiated on two
structurally different action-generation models: a Dita-based \cite{dita}
diffusion model and a Qwen3-VL-2B-based model \cite{qwen3vl} equipped with a
flow-matching action expert. In both implementations, OC-VLA++ does
not introduce additional network modules. The original and synthesized
views are processed by the same model with fully shared parameters,
while the proposed paired-view supervision is applied only through the
training objectives. The model structures are shown in Figure~\ref{fig:model_struct}.

\paragraph{Dita-based model.}
We retain the original Dita~\cite{dita} architecture used in
OC-VLA~\cite{OC-VLA}. Dita directly models continuous action chunks
with a diffusion transformer and does not rely on a pretrained
vision-language backbone. Given an RGB observation, a language
instruction, and the robot proprioceptive state, the corresponding
conditioning features are provided to the diffusion transformer
together with a noisy action chunk. The model predicts the diffusion
noise for all action positions in parallel, from which the clean
continuous action chunk is recovered during denoising.

Following OC-VLA, the action targets are represented in the coordinate
frame of the observing camera rather than the robot base frame. We
first pretrain Dita on DROID~\cite{droid} and subsequently fine-tune it
on each target real-robot or simulation dataset. OC-VLA++ retains the
same conditioning modules, diffusion transformer, and action decoder;
only the synthesized-view generative objective and cross-view
action-equivariance objective are added during fine-tuning.

\paragraph{Qwen3-VL-2B action-expert model.}
Our second implementation follows the action-expert formulation of
\(\pi_0\)~\cite{pi0}. A pretrained Qwen3-VL-2B
backbone~\cite{qwen3vl} encodes the RGB observation and language
instruction, while a dedicated action expert models the robot state
and continuous action chunk. The multimodal context produced by the
vision-language backbone conditions the action expert, which predicts
the flow velocity along a linear probability path. The resulting
velocity field is integrated at inference time to generate the final
continuous action sequence.

The robot state and noisy action chunk are projected into the action
expert's hidden space before being processed by its transformer
blocks. Unlike the Dita-based model, which learns its multimodal
conditioning from robot data, this model inherits image-conditioned
semantic representations from the pretrained Qwen3-VL-2B backbone.
We therefore fine-tune it directly on each target dataset without an
intermediate DROID pretraining stage. As in the Dita implementation,
the model predicts camera-space actions and its architecture remains
unchanged when equipped with OC-VLA++.

\par\medskip
\noindent
During OC-VLA++ training, the original and synthesized observations
are evaluated in separate forward passes using the same network
parameters and the same sampled generative timestep. Each observation
is supervised by its own camera-space action target, while their
recovered action predictions are aligned in the shared robot frame.
At inference time, only the observed RGB image, language instruction,
and robot state are provided to the model; the synthesized-view branch
and monocular geometry estimator are not used.

\subsection{Optimization Details}

All models are optimized using AdamW \cite{adamw} with
\(\beta_1=0.9\), \(\beta_2=0.95\), and a weight decay of \(0.05\).
The learning rate is linearly warmed up for one epoch and subsequently
decayed to \(1\%\) of its initial value following a half-cycle cosine
schedule. Within each model family, the robot-base-frame baseline,
OC-VLA \cite{OC-VLA}, and OC-VLA++ use identical model initialization, data
sampling, optimization settings, and training duration.

\paragraph{Dita-based models.}
Following OC-VLA~\cite{OC-VLA}, Dita is first pretrained on
DROID~\cite{droid} and subsequently adapted to each target dataset.
The pretrained visual encoder is optimized with a learning rate of
\(1\times10^{-5}\), while all remaining model components, including
the Q-Former and causal transformer, are optimized with a learning
rate of \(1\times10^{-4}\).

For the real-robot experiments, the model is fine-tuned for 20,000
optimization steps with a global batch size of 512. For the
ManiSkill2 experiments, the model is trained for 30,000 optimization
steps with a global batch size of 2,048. The diffusion process uses
100 DDPM timesteps during training, and inference is performed using
10-step DDIM sampling.

\paragraph{Qwen3-VL-2B action-expert models.}
The Qwen3-VL-2B-based models are initialized from the pretrained
vision-language backbone and fine-tuned directly on each target
dataset without intermediate DROID pretraining. All parameters of the
Qwen3-VL-2B \cite{qwen3vl} backbone and action expert participate in training and are
jointly optimized using a uniform learning rate of
\(1\times10^{-4}\). We use the same AdamW configuration, warmup
strategy, and cosine learning-rate schedule as for the Dita-based
models.

For the real-robot experiments, the model is trained for 20,000
optimization steps with a global batch size of 512. During flow-matching training, the
timestep is sampled directly from
\(\tau\sim\operatorname{Beta}(1.5,1.0)\). At inference time, the probability flow is solved
using 10 integration steps.

\paragraph{OC-VLA++ objectives.}
We set the weight of the synthesized-view generative objective to
\(\lambda_{\mathrm{syn}}=0.5\) and the weight of the cross-view
action-equivariance objective to \(\lambda_{\mathrm{eq}}=0.5\).
Within the action-equivariance objective, the translation, rotation,
and gripper discrepancies are weighted by
\(\lambda_p=1.0\), \(\lambda_R=0.2\), and
\(\lambda_g=0.1\), respectively. The corresponding Huber parameters
are set to
\(\beta_p=0.01\), \(\beta_R=0.05\), and
\(\beta_g=1.0\). To limit the influence of large discrepancies caused
by synthesis artifacts, we cap the translation and rotation terms
using \(\kappa_p=0.5\) and \(\kappa_R=0.5\), respectively.

For each paired training example, the original and synthesized views
use the same sampled generative timestep. At every optimization step,
one synthesized view is sampled uniformly from the corresponding
offline view pool. All variants within the same model family use the
same optimization configuration and differ only in their action
representation and enabled objective terms.

\begin{table}[t]
\centering
\setlength{\tabcolsep}{8pt}
\begin{tabular}{ccc}
\toprule
\(\lambda_{\mathrm{syn}}\) &
\(\lambda_{\mathrm{eq}}\) &
Average Success Rate (\%) \\
\midrule
0.00 & 0.00 & 52.4 \\
0.50 & 0.00 & 53.4 \\
0.00 & 0.50 & 54.2 \\
\midrule
0.25 & 0.50 & 55.8 \\
\textbf{0.50} & \textbf{0.50} & \textbf{56.8} \\
1.00 & 0.50 & 56.0 \\
\midrule
0.50 & 0.25 & 55.6 \\
0.50 & 1.00 & 55.4 \\
\bottomrule
\end{tabular}
\caption{
Sensitivity analysis of the synthesized-view supervision weight
\(\lambda_{\mathrm{syn}}\) and action-equivariance weight
\(\lambda_{\mathrm{eq}}\) on ManiSkill2. We vary one parameter at a
time while fixing the other to its default value of \(0.5\).
Each configuration is evaluated over 100 episodes on each of the five
tasks.
}
\label{tab:loss_weight_sensitivity}
\end{table}

\subsection{Sensitivity Analysis of Loss Weights}
\label{app:loss_weight_sensitivity}

We further examine the sensitivity of OC-VLA++ to the weights of the
synthesized-view generative objective and the cross-view
action-equivariance objective. As shown in
Table~\ref{tab:loss_weight_sensitivity}, OC-VLA++ performs consistently
well across a moderate range of objective weights. Reducing either
\(\lambda_{\mathrm{syn}}\) or \(\lambda_{\mathrm{eq}}\) to \(0.25\)
leads to a modest performance drop, indicating that both
synthesized-view supervision and cross-view alignment contribute to
effective paired-view training.

Increasing either weight to \(1.0\) also slightly reduces performance.
This may reflect a stronger influence of reconstruction artifacts when
the synthesized-view objective is over-weighted, or an overly
restrictive consistency constraint when the equivariance objective is
assigned excessive weight. The default setting
\(\lambda_{\mathrm{syn}}=\lambda_{\mathrm{eq}}=0.5\) achieves the
highest average success rate. Meanwhile, the relatively small
variation among neighboring configurations indicates that OC-VLA++ is
not highly sensitive to the precise choice of these objective weights.

Overall, these results suggest that the two objectives remain
complementary over a reasonably broad range of relative weights,
rather than requiring a narrowly tuned balance to yield improvements.

\subsection{Qualitative Results of ManiSkill2}

\begin{figure}[!htbp]
\centering
\includegraphics[width=\linewidth]{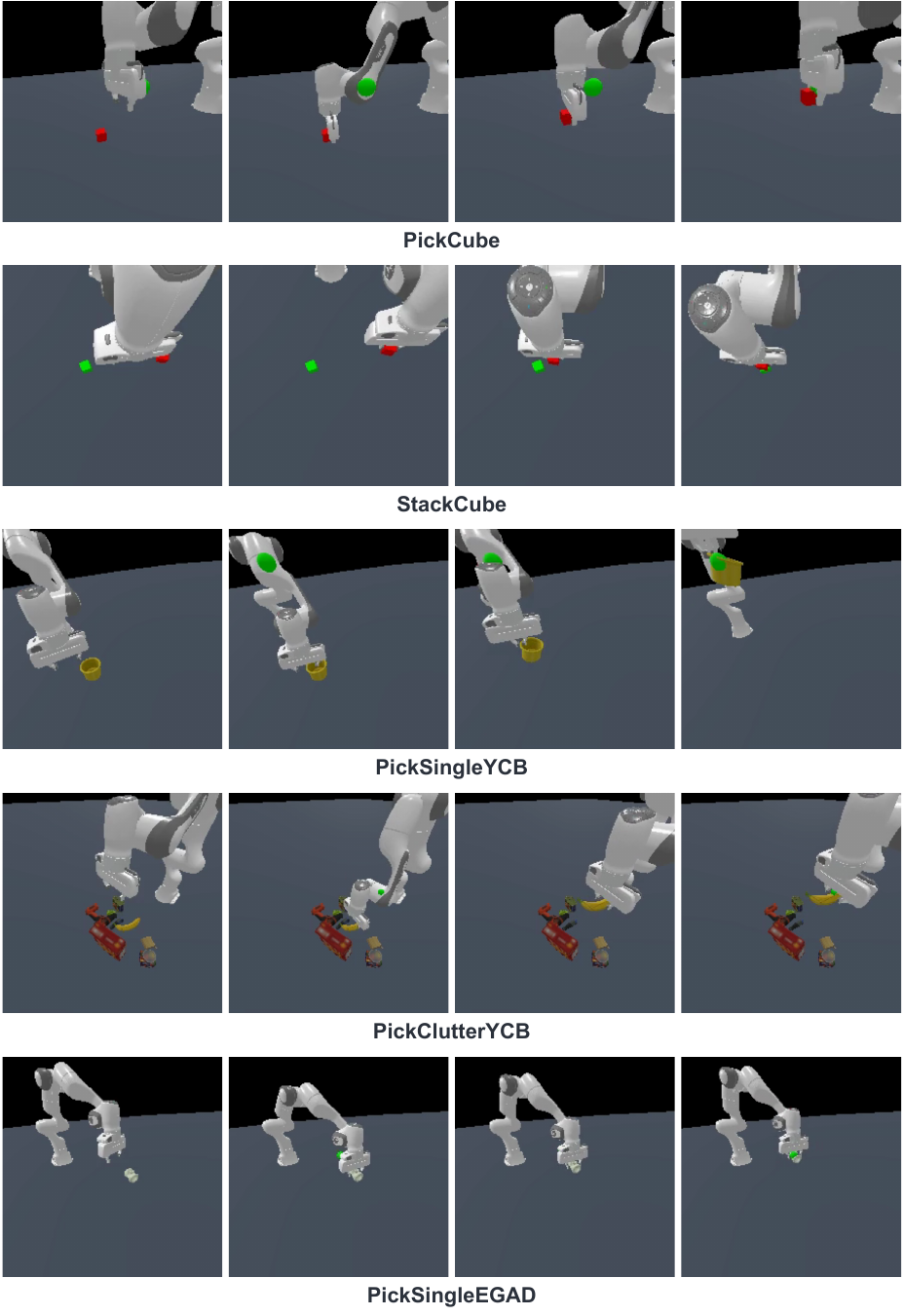}
% \fbox{\parbox[c][1.75in][c]{0.94\textwidth}{
\centering

\caption{
Qualitative results on ManiSkill2 \cite{maniskill2}. Each row corresponds to one task,
while the columns show representative successful rollouts under
different camera viewpoints. OC-VLA++ consistently produces valid
manipulation behaviors across diverse object configurations and
viewing conditions.
}
\label{fig:mani_qual}
\end{figure}

\begin{figure*}[!htbp]
\centering
\includegraphics[width=\linewidth]{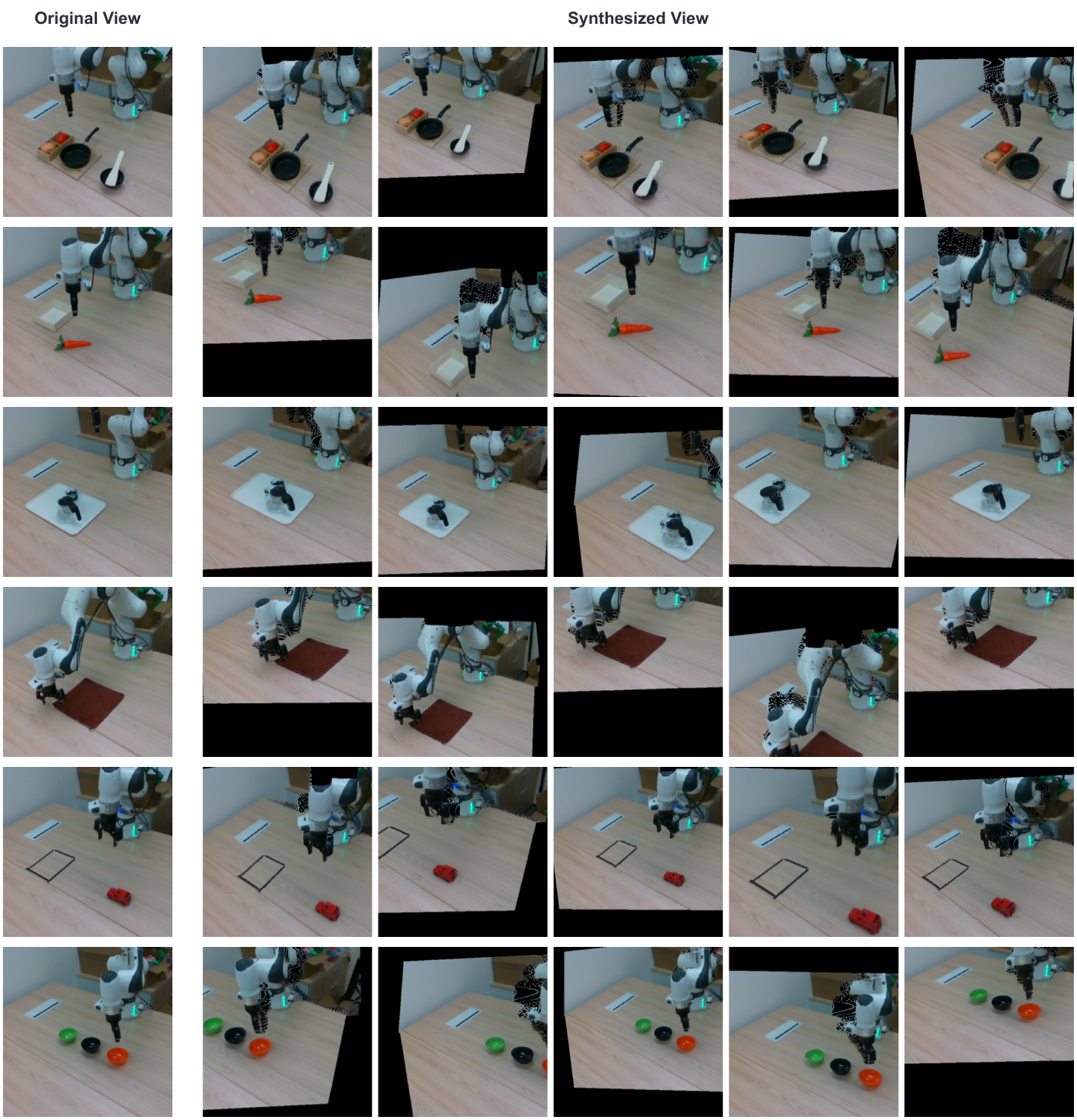}
% \fbox{\parbox[c][1.75in][c]{0.94\textwidth}{
\centering

\caption{
Examples of geometry-guided synthesized views for real-robot
training data with MoGe-2 \cite{moge2}. The first column shows the original observation, and
the remaining columns show nearby views generated from the same
monocular reconstruction under different sampled camera
perturbations. Black regions correspond to pixels without valid
projected geometry. Despite partial disocclusions and reprojection
artifacts, the synthesized observations preserve the task-relevant
scene content and provide geometrically related paired-view
supervision.
}
\label{fig:moge2_demo}
\end{figure*}
\begin{table*}[!t]
\centering
\setlength{\tabcolsep}{4.5pt}
\renewcommand{\arraystretch}{1.18}

\begin{tabular*}{0.94\textwidth}{
@{\extracolsep{\fill}}
c c c c c
@{}
}
\toprule
\textbf{Position} &
\textbf{Translation \(\mathbf{t}\)} &
\textbf{Quaternion \(\mathbf{q}\)} &
\textbf{\(\Delta t\) (cm)} &
\textbf{\(\Delta R\) (\(^\circ\))} \\
\midrule

0 &
\([1.14104,\;0.64038,\;0.87200]\) &
\([0.41022,\;0.81228,\;-0.39164,\;-0.15088]\) &
0.00 &
0.00 \\

1 &
\([1.12895,\;0.74467,\;0.83532]\) &
\([0.41803,\;0.78702,\;-0.40545,\;-0.18407]\) &
11.12 &
5.07 \\

2 &
\([0.97982,\;0.77113,\;0.79215]\) &
\([0.30377,\;0.83719,\;-0.43853,\;-0.12209]\) &
22.24 &
14.02 \\

3 &
\([0.88241,\;0.80146,\;0.86221]\) &
\([0.22221,\;0.89222,\;-0.40162,\;-0.13082]\) &
30.48 &
23.40 \\
\bottomrule
\end{tabular*}

\caption{
Calibrated camera poses used in the real-robot displacement
evaluation. Translations are expressed in meters in the robot base
frame, and quaternions follow the \(xyzw\) convention. Offsets are
measured relative to Position~0.
}
\label{tab:camera_extrinsics}
\end{table*}

Figure~\ref{fig:mani_qual} presents representative
successful episodes of OC-VLA++ on the five ManiSkill2 \cite{maniskill2} tasks. Each row
corresponds to one task, and the columns show observations captured
from different camera viewpoints. Despite substantial changes in
object appearance, spatial layout, and robot projection across views,
the model consistently predicts executable actions that complete the
corresponding manipulation task. These qualitative results complement
the quantitative evaluation by illustrating the viewpoint robustness
of OC-VLA++ across both rigid-object manipulation and cluttered-scene
interaction.

\subsection{Paired-View Generation Details}

We provide additional implementation details for constructing the
paired views used by OC-VLA++. For real-robot data, synthesized views
are generated offline from monocular geometry, whereas ManiSkill2
provides exact paired observations through simulator re-rendering.

\paragraph{Real-robot view synthesis.}
For each real robot RGB observation, we apply a frozen MoGe-2
ViT-Large model with normal prediction~\cite{moge2}.
We use the calibrated physical-camera intrinsics for geometry recovery
and subsequent reprojection, rather than relying on the camera
intrinsics estimated by MoGe-2. Although the model predicts point,
depth, normal, and validity maps, our synthesis pipeline uses only the
metric-scale point map and binary validity mask. The surface-normal map
is not used.

For every original observation, we construct an offline view bank
containing 10 nearby synthesized views. The sampled camera
perturbations follow the ranges specified in the main paper, with
translation bounded by \(15\,\mathrm{cm}\) and rotation bounded by
\(10^\circ\). Valid reconstructed points are transformed into the
sampled target-camera frame and projected onto the image plane using
the calibrated intrinsics. When multiple points project to the same
pixel, the point with the smallest positive target-camera depth is
retained.

Pixels without valid projected geometry are assigned zero RGB values
and therefore appear black in the synthesized image. The corresponding
validity mask is used only during offline generation and is not
provided to the action-generation model. Views with insufficient
valid-pixel coverage are discarded, and camera perturbations are
resampled until 10 valid views have been retained for each original
observation. The same filtering rule is applied across all real-robot
tasks without task-specific adjustment.

Representative view banks are shown in
Figure~\ref{fig:moge2_demo}. The examples illustrate
that local viewpoint changes preserve the principal objects, robot
configuration, and task-relevant spatial relationships, while regions
newly exposed by camera motion remain unreconstructed. These views are
not intended to provide photorealistic novel-view synthesis; instead,
they serve as geometrically related, imperfect but useful training
observations.

\paragraph{Ground-truth paired views in simulation.}
For ManiSkill2~\cite{maniskill2}, we follow the data-generation
protocol of OC-VLA~\cite{OC-VLA}. For each trajectory, 20 camera poses
are sampled from the same pool of 300K candidate poses. Paired
observations are constructed by preserving the complete simulator
state and re-rendering it from an additional sampled camera pose.
Consequently, the paired observations share exactly the same robot and
object states, while their RGB observations and camera extrinsics are
obtained directly from the simulator without monocular reconstruction
or reprojection artifacts.

\begin{table*}[t]
\centering

\setlength{\tabcolsep}{2.8pt}
\renewcommand{\arraystretch}{1.08}

{
\def\pct#1{#1\%}

\begin{tabular*}{\textwidth}{
@{\extracolsep{\fill}}
c c l r r r r r r r
@{}
}
\toprule

\multirow{2}{*}{\textbf{Pos.}} &
\multirow{2}{*}{\textbf{Backbone}} &
\multicolumn{1}{c}{\multirow{2}{*}{\textbf{Setting}}} &
\multicolumn{1}{c}{\textbf{Pick}} &
\multicolumn{1}{c}{\textbf{Pour}} &
\multicolumn{1}{c}{\textbf{Fold}} &
\multicolumn{1}{c}{\textbf{Stack}} &
\multicolumn{1}{c}{\textbf{Push}} &
\multicolumn{1}{c}{\textbf{Cook}} &
\multicolumn{1}{c}{\multirow{2}{*}{\textbf{Avg.}}}
\\

& & &
\multicolumn{1}{c}{\textbf{Carrot}} &
\multicolumn{1}{c}{\textbf{Water}} &
\multicolumn{1}{c}{\textbf{Towel}} &
\multicolumn{1}{c}{\textbf{Bowls}} &
\multicolumn{1}{c}{\textbf{Cars}} &
\multicolumn{1}{c}{\textbf{Potato}} &
\\

\midrule

\multirow{6}{*}{0}
& \multirow{3}{*}{Dita}
& Baseline
& \pct{75.0} & \pct{55.0} & \pct{85.0}
& \pct{45.0} & \pct{50.0} & \pct{60.0}
& \pct{61.7} \\
&
& OC-VLA
& \pct{70.0} & \pct{60.0} & \pct{95.0}
& \pct{40.0} & \pct{70.0} & \pct{75.0}
& \textbf{\pct{68.3}} \\
&
& OC-VLA++
& \pct{80.0} & \pct{50.0} & \pct{95.0}
& \pct{50.0} & \pct{60.0} & \pct{75.0}
& \textbf{\pct{68.3}} \\
\cmidrule(lr){2-10}
&
\multirow{3}{*}{Qwen3-VL}
& Baseline
& \pct{70.0} & \pct{40.0} & \pct{65.0}
& \pct{30.0} & \pct{50.0} & \pct{60.0}
& \pct{52.5} \\
&
& OC-VLA
& \pct{75.0} & \pct{50.0} & \pct{80.0}
& \pct{35.0} & \pct{60.0} & \pct{60.0}
& \textbf{\pct{60.0}} \\
&
& OC-VLA++
& \pct{75.0} & \pct{45.0} & \pct{75.0}
& \pct{45.0} & \pct{55.0} & \pct{60.0}
& \pct{59.2} \\
\midrule

\multirow{6}{*}{1}
& \multirow{3}{*}{Dita}
& Baseline
& \pct{75.0} & \pct{40.0} & \pct{70.0}
& \pct{25.0} & \pct{45.0} & \pct{50.0}
& \pct{50.8} \\
&
& OC-VLA
& \pct{70.0} & \pct{50.0} & \pct{80.0}
& \pct{30.0} & \pct{60.0} & \pct{60.0}
& \pct{58.3} \\
&
& OC-VLA++
& \pct{80.0} & \pct{45.0} & \pct{85.0}
& \pct{40.0} & \pct{55.0} & \pct{65.0}
& \textbf{\pct{61.7}} \\
\cmidrule(lr){2-10}
&
\multirow{3}{*}{Qwen3-VL}
& Baseline
& \pct{65.0} & \pct{25.0} & \pct{50.0}
& \pct{20.0} & \pct{45.0} & \pct{50.0}
& \pct{42.5} \\
&
& OC-VLA
& \pct{80.0} & \pct{40.0} & \pct{60.0}
& \pct{35.0} & \pct{60.0} & \pct{55.0}
& \pct{55.0} \\
&
& OC-VLA++
& \pct{80.0} & \pct{45.0} & \pct{60.0}
& \pct{35.0} & \pct{60.0} & \pct{60.0}
& \textbf{\pct{56.7}} \\
\midrule

\multirow{6}{*}{2}
& \multirow{3}{*}{Dita}
& Baseline
& \pct{65.0} & \pct{30.0} & \pct{50.0}
& \pct{20.0} & \pct{45.0} & \pct{40.0}
& \pct{41.7} \\
&
& OC-VLA
& \pct{65.0} & \pct{35.0} & \pct{65.0}
& \pct{30.0} & \pct{50.0} & \pct{55.0}
& \pct{50.0} \\
&
& OC-VLA++
& \pct{70.0} & \pct{45.0} & \pct{70.0}
& \pct{35.0} & \pct{55.0} & \pct{55.0}
& \textbf{\pct{55.0}} \\
\cmidrule(lr){2-10}
&
\multirow{3}{*}{Qwen3-VL}
& Baseline
& \pct{55.0} & \pct{20.0} & \pct{45.0}
& \pct{15.0} & \pct{45.0} & \pct{45.0}
& \pct{37.5} \\
&
& OC-VLA
& \pct{70.0} & \pct{30.0} & \pct{50.0}
& \pct{25.0} & \pct{50.0} & \pct{50.0}
& \pct{45.8} \\
&
& OC-VLA++
& \pct{75.0} & \pct{40.0} & \pct{55.0}
& \pct{30.0} & \pct{55.0} & \pct{60.0}
& \textbf{\pct{52.5}} \\
\midrule

\multirow{6}{*}{3}
& \multirow{3}{*}{Dita}
& Baseline
& \pct{50.0} & \pct{20.0} & \pct{35.0}
& \pct{10.0} & \pct{40.0} & \pct{35.0}
& \pct{31.7} \\
&
& OC-VLA
& \pct{60.0} & \pct{30.0} & \pct{50.0}
& \pct{20.0} & \pct{45.0} & \pct{40.0}
& \pct{40.8} \\
&
& OC-VLA++
& \pct{65.0} & \pct{40.0} & \pct{60.0}
& \pct{30.0} & \pct{55.0} & \pct{40.0}
& \textbf{\pct{48.3}} \\
\cmidrule(lr){2-10}
&
\multirow{3}{*}{Qwen3-VL}
& Baseline
& \pct{45.0} & \pct{15.0} & \pct{35.0}
& \pct{10.0} & \pct{40.0} & \pct{40.0}
& \pct{30.8} \\
&
& OC-VLA
& \pct{55.0} & \pct{25.0} & \pct{40.0}
& \pct{20.0} & \pct{45.0} & \pct{40.0}
& \pct{37.5} \\
&
& OC-VLA++
& \pct{65.0} & \pct{35.0} & \pct{45.0}
& \pct{25.0} & \pct{45.0} & \pct{45.0}
& \textbf{\pct{43.3}} \\
\bottomrule
\end{tabular*}
}

\caption{
Per-task success rates (\%) under increasing camera displacement.
Each entry is evaluated over 20 real-robot trials, and the average is
computed across the six tasks. Camera positions correspond to
Table~\ref{tab:camera_extrinsics}. Bold values indicate the best
average within each backbone at each camera position.
}
\label{tab:per_task_camera_displacement}

\end{table*}

\subsection{Detailed Results under Camera Displacement}

Table~\ref{tab:camera_extrinsics} reports the calibrated extrinsics of
the four physical camera configurations used in the displacement
evaluation. Relative to the nominal configuration at Position~0, the
camera undergoes translation offsets of \(11.1\), \(22.2\), and
\(30.5\) cm at Positions~1--3, together with rotation offsets of
\(5.1^\circ\), \(14.0^\circ\), and \(23.4^\circ\), respectively. The
simultaneous increase in translation and rotation produces
progressively stronger changes in object appearance, robot projection,
and spatial relationships in the image. These measurements provide the
camera configurations corresponding to the averaged displacement
curves in the main paper.

Table~\ref{tab:per_task_camera_displacement} gives the complete
per-task success rates. At Position~0, OC-VLA++ preserves performance
under the nominal viewpoint. With Dita \cite{dita}, OC-VLA \cite{OC-VLA} and OC-VLA++ both
achieve an average success rate of \(68.3\%\). With Qwen3-VL \cite{qwen3vl}, the two
methods achieve \(60.0\%\) and \(59.2\%\), respectively, corresponding
to only one successful trial of difference among the 120 evaluations
performed across the six tasks. This result indicates that the
additional paired-view supervision does not introduce a meaningful
trade-off at the original camera configuration.

The difference becomes clearer as the camera moves away from
Position~0. With Dita, OC-VLA++ improves over OC-VLA by \(3.4\),
\(5.0\), and \(7.5\) percentage points at Positions~1--3,
respectively. From Position~0 to Position~3, the average success rate
of OC-VLA decreases from \(68.3\%\) to \(40.8\%\), whereas OC-VLA++
decreases to \(48.3\%\). Thus, OC-VLA++ experiences a \(20.0\)-point
drop, compared with a \(27.5\)-point drop for OC-VLA and a
\(30.0\)-point drop for the robot-base-frame Dita baseline. A similar
pattern is observed with Qwen3-VL. At Positions~2 and~3, OC-VLA++
outperforms OC-VLA by \(6.7\) and \(5.8\) percentage points,
respectively, and its overall decrease from Position~0 to Position~3
is \(15.9\) points, compared with \(22.5\) points for OC-VLA. These
results provide a more direct view of the slower degradation exhibited
by OC-VLA++ under increasingly large viewpoint shifts.

The task-level breakdown further shows that the average improvement is
not dominated by a single task. At Position~3, the Dita-based
OC-VLA++ improves over OC-VLA on five of the six tasks, including gains
of \(10\) percentage points on Pour Water, Fold Towel, Stack Bowls,
and Push Cars. With Qwen3-VL, improvements are observed on Pick
Carrot, Pour Water, Fold Towel, Stack Bowls, and Cook Potato at the
same camera position. Since each task is evaluated over 20 trials, a
\(5\)-point difference corresponds to one additional successful
rollout. The consistent gains across multiple tasks and both model
families therefore suggest that the improved robustness reflects a
general reduction in viewpoint sensitivity rather than an isolated
benefit on a particular manipulation scenario.

\end{document}